%% file: main.tex
\documentclass{article}

\usepackage{hyperref}       
\usepackage{url}            
\usepackage{booktabs}       
\usepackage{amsfonts}       
\usepackage{nicefrac}       
\usepackage{microtype}      
\usepackage{graphicx}
\usepackage{amsmath,amssymb} 
\usepackage{color}
\usepackage{epsfig}
\usepackage{tabularx}
\usepackage{booktabs}
\usepackage{multirow}
\usepackage{array}
\usepackage{xspace}
\usepackage{balance}
\usepackage{enumitem}
\usepackage{mathtools}
\usepackage{rotating}
\usepackage{tabulary}
\usepackage[table,dvipsnames]{xcolor}
\usepackage{subcaption}
\usepackage[export]{adjustbox}
\usepackage{ctable}
\usepackage{diagbox}
\usepackage{makecell}
\usepackage{bbding}
\usepackage{pifont}
\usepackage{stfloats}

\newcommand{\xmark}{\text{\ding{55}}}

\newcommand{\ie}{\textit{i}.\textit{e}.}
\newcommand{\eg}{\textit{e}.\textit{g}.}

\def\name{Q-Former}

\usepackage[accepted]{icml2023}

\icmltitlerunning{BLIP-2: Bootstrapping Language-Image Pre-training with Frozen Image Encoders and Large Language Models}

\begin{document}
\twocolumn[
\icmltitle{BLIP-2: Bootstrapping Language-Image Pre-training \\ with Frozen Image Encoders and Large Language Models}



\icmlsetsymbol{equal}{*}

\begin{icmlauthorlist}
\icmlauthor{Junnan Li}{}
\icmlauthor{Dongxu Li}{}
\icmlauthor{Silvio Savarese}{}
\icmlauthor{Steven Hoi}{}
\\
\icmlauthor{Salesforce Research}{}
\\
\url{https://github.com/salesforce/LAVIS/tree/main/projects/blip2}
\end{icmlauthorlist}


\icmlkeywords{vision-language}

\vskip 0.3in
]




\input{0-abstract}
\input{1-intro}
\input{2-related}

\input{3-method}

\input{4-experiment}
\input{5-limitation}


\bibliographystyle{icml2023}
\bibliography{main}

\input{appendix.tex}

\end{document}

%% file: 0-abstract.tex
\begin{abstract}

The cost of vision-and-language pre-training has become increasingly prohibitive due to end-to-end training of large-scale models.
This paper proposes BLIP-2, a generic and efficient pre-training strategy that bootstraps vision-language pre-training from off-the-shelf frozen pre-trained image encoders and frozen large language models.
BLIP-2 bridges the modality gap with a lightweight Querying Transformer, which is pre-trained in two stages.
The first stage bootstraps vision-language representation learning from a frozen image encoder.
The second stage bootstraps vision-to-language generative learning from a frozen language model.
BLIP-2 achieves state-of-the-art performance on various vision-language tasks,
despite having significantly fewer trainable parameters than existing methods.
For example,
our model outperforms Flamingo80B by 8.7\% on
zero-shot VQAv2 with 54x fewer trainable parameters.
We also demonstrate the model's emerging capabilities of zero-shot image-to-text generation that can follow natural language instructions.
\end{abstract}

%% file: 1-intro.tex
\vspace{-2ex}
\section{Introduction}
\vspace{-0.5ex}
\label{sec:intro}




Vision-language pre-training (VLP) research has witnessed a rapid advancement in the past few years,
where pre-trained models with increasingly larger scale have been developed to continuously push the state-of-the-art on various downstream tasks~\cite{clip,ALBEF,blip,ofa,flamingo,beit3}.
However, most state-of-the-art vision-language models incur a high computation cost during pre-training, due to
end-to-end training using large-scale models and datasets.
\input{figure/teaser.tex}

Vision-language research sits at the intersection between vision and language,
therefore it is naturally expected that vision-language models can harvest from the readily-available unimodal models from the vision and natural language communities.
In this paper, we propose a \textit{generic} and \textit{compute-efficient} VLP method by 
bootstrapping from off-the-shelf pre-trained vision models and language models.
Pre-trained vision models offer high-quality visual representation.
Pre-trained language models, in particular \textit{large language models} (LLMs), offer strong language generation and zero-shot transfer abilities.
To reduce computation cost and counteract the issue of catastrophic forgetting, the unimodal pre-trained models remain frozen during the pre-training.

In order to leverage pre-trained unimodal models for VLP, it is key to facilitate cross-modal alignment.
However, since LLMs have not seen images during their unimodal pre-training, 
freezing them makes vision-language alignment in particular challenging.
In this regard, existing methods (\eg~Frozen~\cite{Frozen}, Flamingo~\cite{flamingo}) resort to an image-to-text generation loss,
which we show is insufficient to bridge the modality gap.

To achieve effective vision-language alignment with frozen unimodal models, we propose a Querying Transformer (\name) pre-trained with a new two-stage pre-training strategy.
As shown in Figure~\ref{fig:teaser},
\name~is a lightweight transformer which employs a set of learnable query vectors to extract visual features from the frozen image encoder.
It acts as an information bottleneck between the frozen image encoder and the frozen LLM,
where it feeds the most useful visual feature for the LLM to output the desired text.
In the first pre-training stage,
we perform vision-language representation learning
which enforces the \name~to learn visual representation most relevant to the text.
In the second pre-training stage,
we perform vision-to-language generative learning by connecting the output of the \name~to a frozen LLM,
and trains the \name~such that its output visual representation can be interpreted by the LLM.

We name our VLP framework as BLIP-2: Bootstrapping Language-Image Pre-training with frozen unimodal models.
The key advantages of BLIP-2 include:
\vspace{-\topsep}
\vspace{-0.5ex}
\begin{itemize} [leftmargin=*]
    \item BLIP-2 effectively leverages both frozen pre-trained image models and language models.
    We bridge the modality gap using a \name~pre-trained in two-stages: representation learning stage and generative learning stage.
    BLIP-2 achieves state-of-the-art performance on various vision-language tasks including visual question answering, image captioning, and image-text retrieval.
    \item
    Powered by LLMs (\eg~OPT~\cite{opt}, FlanT5~\cite{flanT5}), BLIP-2 can be prompted to perform zero-shot image-to-text generation that follows natural language instructions, which enables emerging capabilities such as visual knowledge reasoning, visual conversation, etc. (see Figure~\ref{fig:example} for examples).    
    \item
    Due to the use of frozen unimodal models and a lightweight \name,
    BLIP-2 is more compute-efficient than exisiting state-of-the-arts.
    For example, BLIP-2 outperforms Flamingo~\cite{flamingo} by 8.7\% on zero-shot VQAv2, while using 54$\times$ fewer trainable parameters.
    Furthermore,
    our results show that BLIP-2 is a generic method that can harvest more advanced unimodal models for better VLP performance.
\end{itemize}


%% file: figure/teaser.tex


\begin{figure}[!t]
\centering
\includegraphics[width=1\columnwidth]{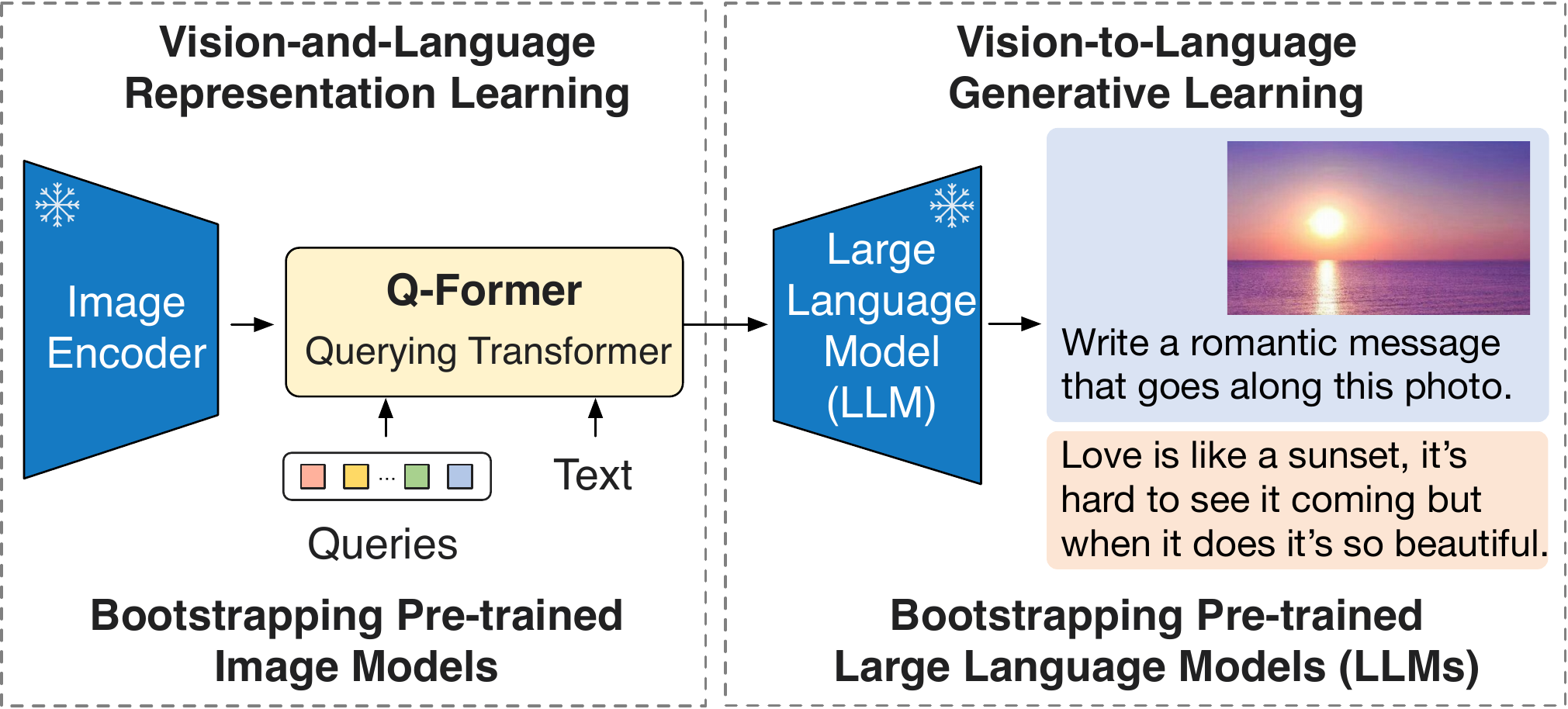}
\vspace{-3.5ex}
\caption{Overview of BLIP-2's framework.
We pre-train a lightweight Querying Transformer following a two-stage strategy to bridge the modality gap.
The first stage bootstraps vision-language representation learning from a frozen image encoder.
The second stage bootstraps vision-to-language generative learning from a frozen LLM,
which enables zero-shot instructed image-to-text generation (see Figure~\ref{fig:example} for more examples).}
\vspace{-2.5ex}
\label{fig:teaser}
\end{figure}

%% file: 2-related.tex
\vspace{-1.5ex}
\section{Related Work}
\vspace{-0.5ex}
\label{sec:related}

\subsection{End-to-end Vision-Language Pre-training}
\vspace{-0.5ex}

Vision-language pre-training aims to learn multimodal foundation models with improved performance on various vision-and-language tasks.
Depending on the downstream task, different model architectures have been proposed, including the dual-encoder architecture~\cite{clip,align}, the fusion-encoder architecture~\cite{LXMERT,ALBEF}, the encoder-decoder architecture~\cite{VL_T5,simvlm,pali},
and more recently, the unified transformer architecture~\cite{blip,beit3}.
Various pre-training objectives have also been proposed over the years,
and have progressively converged to a few time-tested ones: image-text contrastive learning~\cite{clip,filip,ALBEF,blip},
image-text matching~\cite{ALBEF,blip,VLMo},
and (masked) language modeling~\cite{ALBEF,blip,coca,beit3}.

Most VLP methods perform end-to-end pre-training using large-scale image-text pair datasets.
As the model size keeps increasing, the pre-training can incur an extremely high computation cost.
Moreover, it is inflexible for end-to-end pre-trained models to leverage readily-available unimodal pre-trained models,
such as LLMs~\cite{gpt3,opt,flanT5}.

\vspace{-0.5ex}
\subsection{Modular Vision-Language Pre-training}
\vspace{-0.5ex}
More similar to us are methods that leverage off-the-shelf pre-trained models and keep them frozen during VLP.
Some methods freeze the image encoder, including the early work which adopts a frozen object detector to extract visual features~\cite{uniter,oscar,vinvl},
and the recent LiT~\cite{LiT} which uses a frozen pre-trained image encoder for CLIP~\cite{clip} pre-training.
Some methods freeze the language model to use the knowledge from LLMs for vision-to-language generation tasks~\cite{Frozen,flamingo,vgpt,mapl,pnp-vqa,img2prompt}.
The key challenge in using a frozen LLM is to align visual features to the text space.
To achieve this, Frozen~\cite{Frozen} finetunes an image encoder whose outputs are directly used as soft prompts for the LLM.
Flamingo~\cite{flamingo} inserts new cross-attention layers into the LLM to inject visual features, 
and pre-trains the new layers on billions of image-text pairs.
Both methods adopt the language modeling loss,
where the language model generates texts conditioned on the image.

Different from existing methods,
BLIP-2 can effectively and efficiently leverage both frozen image encoders and frozen LLMs for various vision-language tasks,
achieving stronger performance at a lower computation cost.

%% file: 3-method.tex
\vspace{-1.5ex}
\section{Method}
\label{sec:method}
\vspace{-0.5ex}

\input{figure/stage1}

We propose BLIP-2, a new vision-language pre-training method that bootstraps from frozen pre-trained unimodal models.
In order to bridge the modality gap,
we propose a Querying Transformer (\name) pre-trained in two stages:
(1) vision-language representation learning stage with a frozen image encoder and (2) vision-to-language generative learning stage with a frozen LLM.
This section first introduces the model architecture of \name,
and then delineates the two-stage pre-training procedures.

\vspace{-0.5ex}
\subsection{Model Architecture}
\vspace{-0.5ex}
We propose \name~as the trainable module to bridge the gap between a frozen image encoder and a frozen LLM.
It extracts a fixed number of output features from the image encoder,
independent of input image resolution.
As shown in Figure~\ref{fig:stage1},
\name~consists of two transformer submodules that share the same self-attention layers: (1) an image transformer that interacts with the frozen image encoder for visual feature extraction, (2) a text transformer that can function as both a text encoder and a text decoder.
We create a set number of learnable query embeddings as input to the image transformer.
The queries interact with each other through self-attention layers, and interact with frozen image features through cross-attention layers (inserted every other transformer block).
The queries can additionally interact with the text through the same self-attention layers.
Depending on the pre-training task,
we apply different self-attention masks to control query-text interaction.
We initialize \name~with the pre-trained weights of BERT$_\text{base}$~\cite{bert}, whereas the cross-attention layers are randomly initialized. 
In total, \name~contains 188M parameters.
Note that the queries are considered as model parameters.

In our experiments,
we use 32 queries where each query has a dimension of 768 (same as the hidden dimension of the \name).
We use $Z$ to denote the output query representation.
The size of $Z$ ($32\times768$) is much smaller than the size of frozen image features (\eg~$257\times1024$ for ViT-L/14).
This bottleneck architecture works together with our pre-training objectives into forcing the queries to extract visual information that is most relevant to the text.

\vspace{-0.5ex}
\subsection{Bootstrap Vision-Language Representation Learning from a Frozen Image Encoder}
\vspace{-0.5ex}
In the representation learning stage, 
we connect \name~to a frozen image encoder and perform pre-training using image-text pairs.
We aim to train the \name~such that the queries can learn to extract visual representation that is most informative of the text.
Inspired by BLIP~\cite{blip}, we jointly optimize three pre-training objectives that share the same input format and model parameters.
Each objective employs a different attention masking strategy between queries and text to control their interaction (see Figure~\ref{fig:stage1}).


\noindent\textbf{Image-Text Contrastive Learning} (ITC) learns to align image representation and text representation such that their mutual information is maximized.
It achieves so by contrasting the image-text similarity of a positive pair against those of negative pairs.
We align the output query representation $Z$ from the image transformer with the text representation $t$ from the text transformer,
where $t$ is the output embedding of the \texttt{[CLS]} token.
Since $Z$ contains multiple output embeddings (one from each query),
we first compute the pairwise similarity between each query output and $t$, and then select the highest one as the image-text similarity.
To avoid information leak, we employ a unimodal self-attention mask,
where the queries and text are not allowed to see each other.
Due to the use of a frozen image encoder, we can fit more samples per GPU compared to end-to-end methods.
Therefore, we use in-batch negatives instead of the momentum queue in BLIP.

\noindent\textbf{Image-grounded Text Generation} (ITG) loss trains the \name~to generate texts, given input images as the condition.
Since the architecture of \name~does not allow direct interactions between the frozen image encoder and the text tokens,
the information required for generating the text must be first extracted by the queries,
and then passed to the text tokens via self-attention layers.
Therefore, the queries are forced to extract visual features that capture all the information about the text. 
We employ a multimodal causal self-attention mask to control query-text interaction, similar to the one used in UniLM~\cite{UniLM}.
The queries can attend to each other but not the text tokens.
Each text token can attend to all queries and its previous text tokens.
We also replace the \texttt{[CLS]} token with a new \texttt{[DEC]} token as the first text token to signal the decoding task.

\noindent\textbf{Image-Text Matching} (ITM)
aims to learn fine-grained alignment between image and text representation.
It is a binary classification task where the model is asked to predict whether an image-text pair is positive
(matched) or negative (unmatched).
We use a bi-directional self-attention mask where all queries and texts can attend to each other.
The output query embeddings $Z$ thus capture multimodal information.
We feed each output query embedding into a two-class linear classifier to obtain a logit,
and average the logits across all queries as the output matching score.
We adopt the hard negative mining strategy from~\citet{ALBEF,blip} to create informative negative pairs.


\input{figure/stage2}

\vspace{-0.5ex}
\subsection{Bootstrap Vision-to-Language Generative Learning from a Frozen LLM}
\vspace{-0.5ex}
In the generative pre-training stage,
we connect \name~(with the frozen image encoder attached) to a frozen LLM to harvest the LLM's generative language capability.
As shown in Figure~\ref{fig:stage2},
we use a fully-connected (FC) layer to linearly project the output query embeddings $Z$ into the same dimension as the text embedding of the LLM. 
The projected query embeddings are then prepended to the input text embeddings.
They function as \textit{soft visual prompts} that condition the LLM on visual representation extracted by the \name.
Since the \name~has been pre-trained to extract language-informative visual representation,
it effectively functions as an information bottleneck that feeds the most useful information to the LLM while removing irrelevant visual information.
This reduces the burden of the LLM to learn vision-language alignment,
thus mitigating the catastrophic forgetting problem.

We experiment with two types of LLMs: decoder-based LLMs and encoder-decoder-based LLMs.
For decoder-based LLMs,
we pre-train with the language modeling loss,
where the frozen LLM is tasked to generate the text conditioned on the visual representation from \name.
For encoder-decoder-based LLMs,
we pre-train with the prefix language modeling loss,
where we split a text into two parts.
The prefix text is concatenated with the visual representation as input to the LLM's encoder.
The suffix text is used as the generation target for the LLM's decoder.

\vspace{-0.5ex}
\subsection{Model Pre-training}
\vspace{-0.5ex}
\noindent\textbf{Pre-training data.}
We use the same pre-training dataset as BLIP with 129M images in total,
including COCO~\cite{coco}, Visual Genome~\cite{VG}, CC3M~\cite{CC}, CC12M~\cite{cc12m}, SBU~\cite{sbu}, and 115M images from the LAION400M dataset~\cite{laion}.
We adopt the CapFilt method~\cite{blip} to create synthetic captions for the web images.
Specifically, we generate 10 captions using the BLIP$_\mathrm{large}$ captioning model, and rank the synthetic captions along with the original web caption based on the image-text similarity produced by a CLIP ViT-L/14 model.
We keep top-two captions per image as training data and randomly sample one at each pre-training step.

\noindent\textbf{Pre-trained image encoder and LLM.}
For the frozen image encoder, we explore two state-of-the-art pre-trained vision transformer models: (1) ViT-L/14 from CLIP~\cite{clip} and (2) ViT-g/14 from EVA-CLIP~\cite{eva}.
We remove the last layer of the ViT and uses the second last layer's output features,
which leads to slightly better performance.
For the frozen language model,
we explore the unsupervised-trained OPT model family~\cite{opt} for decoder-based LLMs,
and the instruction-trained FlanT5 model family~\cite{flanT5} for encoder-decoder-based LLMs.

\input{figure/example.tex}

\noindent\textbf{Pre-training settings.}
We pre-train for 250k steps in the first stage and 80k steps in the second stage.
We use a batch size of 2320/1680 for ViT-L/ViT-g in the first stage and a batch size of 1920/1520 for OPT/FlanT5 in the second stage.
During pre-training, we convert the frozen ViTs' and LLMs' parameters into FP16, except for FlanT5 where we use BFloat16.
We found no performance degradation compared to using 32-bit models.
Due to the use of frozen models,
our pre-training is more computational friendly than existing large-scale VLP methods.
For example,
using a single 16-A100(40G) machine,
our largest model with ViT-g and FlanT5-XXL requires less than 6 days for the first stage and less than 3 days for the second stage.

The same set of pre-training hyper-parameters are used for all models.
We use the AdamW~\cite{adamw} optimizer with $\beta_1=0.9$, $\beta_1=0.98$, and a weight decay of 0.05.
We use a cosine learning rate decay with a peak learning rate of 1e-4 and a linear warmup of 2k steps. 
The minimum learning rate at the second stage is 5e-5.
We use images of size 224$\times$224, augmented with random resized cropping and horizontal flipping.

%% file: figure/stage1.tex




\begin{figure*}[!t]
  \includegraphics[width=\textwidth]{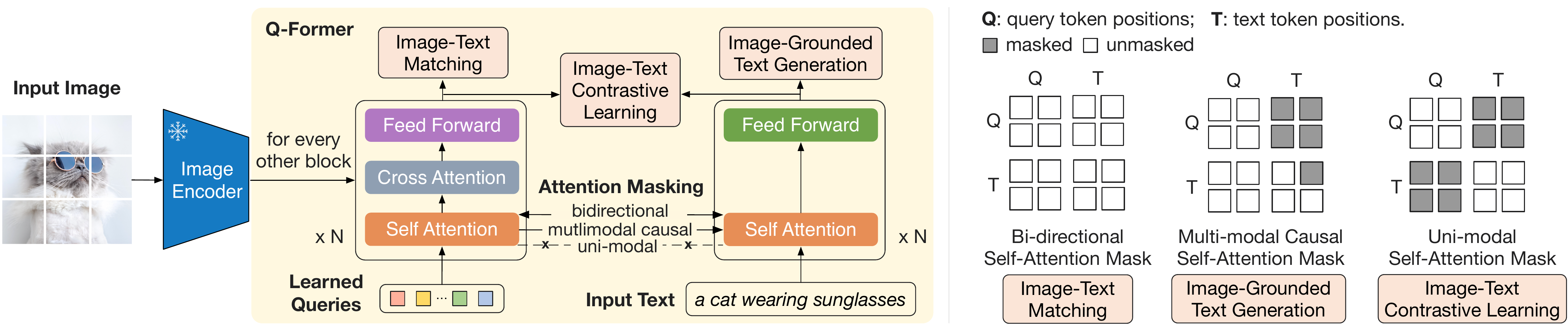}
\vspace{-3.5ex}
\caption{(\textbf{Left}) Model architecture of \name~and BLIP-2's first-stage vision-language representation learning objectives. We jointly optimize three objectives which enforce the queries (a set of learnable embeddings) to extract visual representation most relevant to the text. (\textbf{Right}) The self-attention masking strategy for each objective to control query-text interaction.}
\vspace{-1.5ex}
\label{fig:stage1}
\end{figure*}

%% file: figure/stage2.tex


\begin{figure*}[!t]
\centering
  \includegraphics[width=0.9\textwidth]{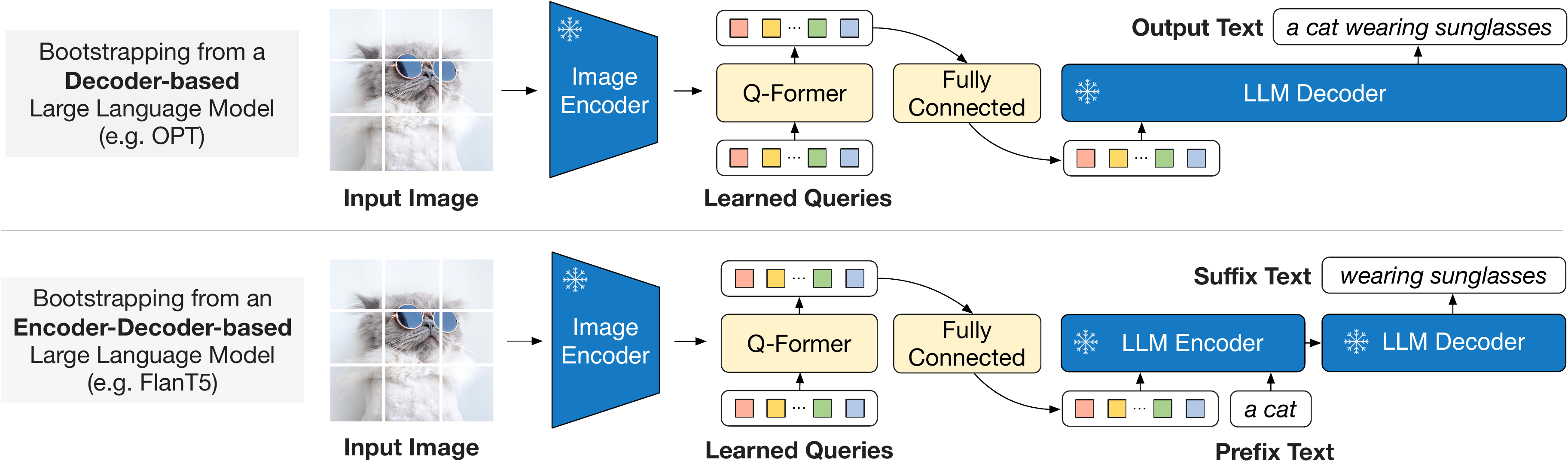}
\vspace{-2ex}
\caption{BLIP-2's second-stage vision-to-language generative pre-training, which bootstraps from frozen large language models (LLMs). (\textbf{Top}) Bootstrapping a decoder-based LLM (e.g. OPT). (\textbf{Bottom}) Bootstrapping an encoder-decoder-based LLM (e.g. FlanT5). The fully-connected layer adapts from the output dimension of the Q-Former to the input dimension of the chosen LLM.}
\vspace{-1.5ex}
\label{fig:stage2}
\end{figure*}



%% file: figure/example.tex
\begin{figure*}[!t]
\centering
  \includegraphics[width=\textwidth]{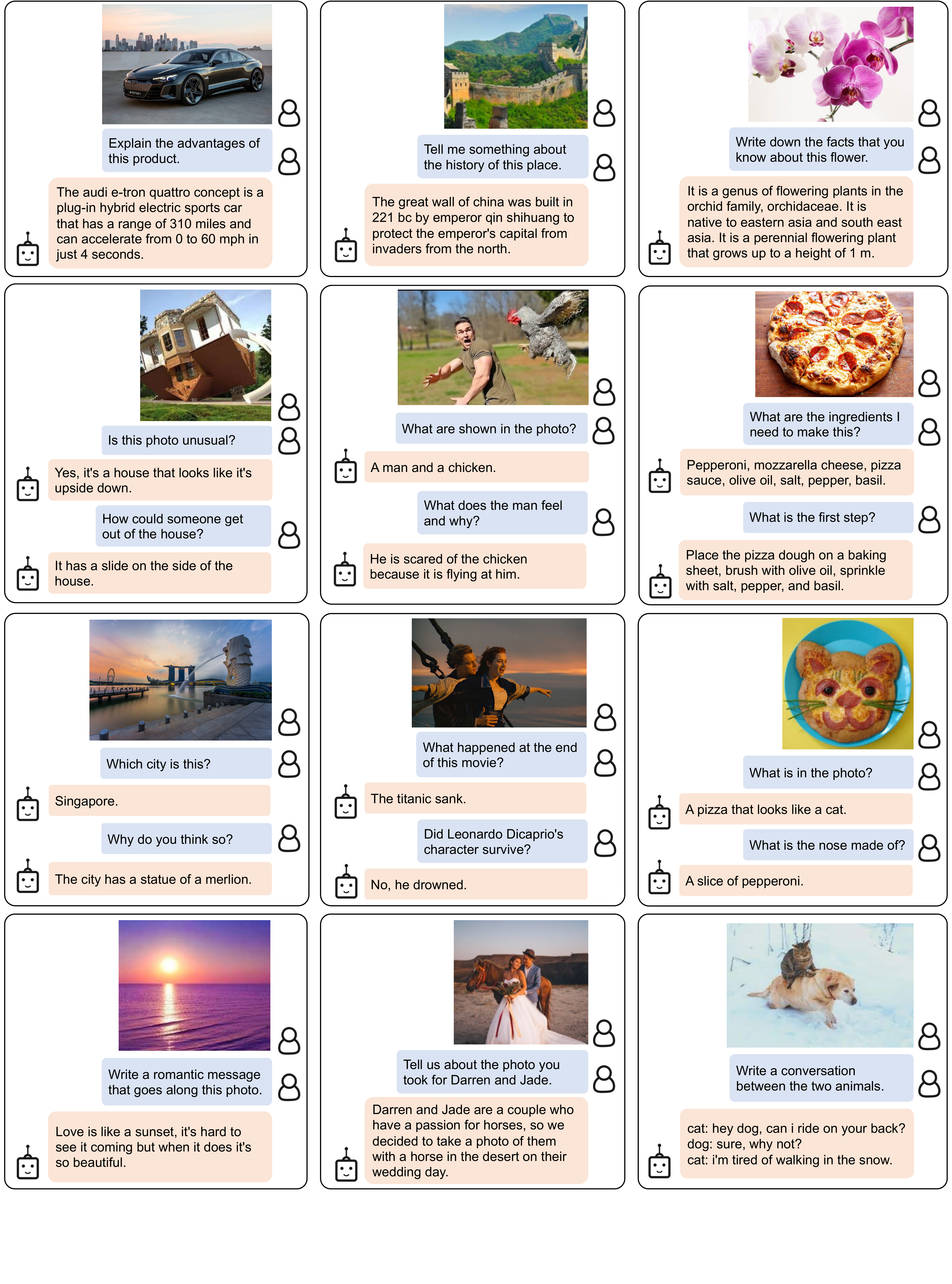}
\vspace{-15ex}
\caption{Selected examples of \textbf{instructed zero-shot image-to-text generation} using a BLIP-2 model w/ ViT-g and FlanT5$_\text{XXL}$,
where it shows a wide range of capabilities including
visual conversation, visual knowledge reasoning, visual commensense reasoning,
storytelling, personalized image-to-text generation, etc.
}
\label{fig:example}
\end{figure*}

%% file: 4-experiment.tex
\vspace{-1ex}
\section{Experiment}
\vspace{-0.5ex}
\label{sec:experiment}

\input{table/zeroshot}
\input{table/vqa_zeroshot}

Table 1 provides an overview of the performance of BLIP-2 on various zero-shot vision-language tasks.
Compared to previous state-of-the-art models,
BLIP-2 achieves improved performance while requiring substantially fewer number of trainable parameters during vision-language pre-training.

\vspace{-0.5ex}
\subsection{Instructed Zero-shot Image-to-Text Generation}
\vspace{-0.5ex}
BLIP-2 effectively enables a LLM to understand images while preserving its capability in following text prompts,
which allows us to control image-to-text generation with instructions.
We simply append the text prompt after the visual prompt as input to the LLM.
Figure~\ref{fig:example} shows examples to demonstrate a wide range of zero-shot image-to-text capabilities including visual knowledge reasoning, visual commensense reasoning,
visual conversation,
personalized image-to-text generation, etc.

\textbf{Zero-shot VQA}.
We perform quantitative evaluation on the zero-shot visual question answering task.
For OPT models, we use the prompt ``Question: {\small\{\}} Answer:''.
For FlanT5 models, we use the prompt ``Question: {\small\{\}} Short answer:''.
During generation,
we use beam search with a beam width of 5.
We also set the length-penalty to -1 which encourages shorter answers that align better with human annotation.

As shown in Table~\ref{tbl:vqa_zeroshot}.
BLIP-2 achieves state-of-the-art result on the VQAv2~\cite{VQA2} and GQA~\cite{GQA} datasets.
It outperforms Flamingo80B by 8.7\% on VQAv2, despite having 54x fewer trainable parameters.
On the OK-VQA~\cite{okvqa} dataset,
BLIP-2 comes secondary to Flamingo80B.
We hypothesis that this is because OK-VQA focuses more on open-world knowledge than visual understanding,
and the 70B Chinchilla~\cite{chinchilla} language model from Flamingo80B possesses more knowledge than the 11B FlanT5$_\text{XXL}$.

We make a promising observation from Table~\ref{tbl:vqa_zeroshot}: 
\textbf{a stronger image encoder or a stronger LLM both lead to better performance.}
This observation is supported by several facts:
(1) ViT-g outperforms ViT-L for both OPT and FlanT5.
(2) Within the same LLM family, larger models outperform smaller ones.
(3) FlanT5, an instruction-tuned LLM, outperforms the unsupervised-trained OPT on VQA. 
This observation validates BLIP-2 as a \textbf{generic vision-language pre-training method} that can efficiently harvest the rapid advances in vision and natural language communities.  

\noindent\textbf{Effect of Vision-Language Representation Learning.}
\vspace{-0.5ex}
\input{figure/qformer_effect.tex}

The first-stage representation learning pre-trains the \name~to learn visual features relevant to the text,
which reduces the burden of the LLM to learn vision-language alignment.
Without the representation learning stage,
\name~relies solely on the vision-to-language generative learning to bridge the modality gap, which is similar to the Perceiver Resampler in Flamingo.
Figure~\ref{fig:qformer_effect} shows the effect of representation learning on generative learning.
Without representation learning,
both types of LLMs give substantially lower performance on zero-shot VQA.
In particular, OPT suffers from catastrophic forgetting where performance drastically degrades as training proceeds.

\input{table/caption}
\input{table/vqa_finetune}

\subsection{Image Captioning}

We finetune BLIP-2 models for the image captioning task,
which asks the model to generate a text description for the image's visual content.
We use the prompt ``a photo of'' as an initial input to the LLM and trains the model to generate the caption with the language modeling loss.
We keep the LLM frozen during finetuning, 
and updates the parameters of the \name~together with the image encoder.
We experiment with ViT-g and various LLMs.
Detailed hyperparameters can be found in the appendix.
We perform finetuning on COCO,
and evaluate on both COCO test set and zero-shot transfer to NoCaps~\cite{nocaps} validation set.

The results are shown in Table~\ref{tbl:caption}.
BLIP-2 achieves state-of-the-art performance with significant improvement on NoCaps over existing methods,
demonstrating strong generalization ability to out-domain images.

\subsection{Visual Question Answering}

Given annotated VQA data,
we finetune the parameters of the \name~and the image encoder while keeping the LLM frozen. 
We finetune with the open-ended answer generation loss,
where the LLM receives \name's output and the question as input,
and is asked to generate the answer.
In order to extract image features that are more relevant to the question,
we additionally condition \name~on the question.
Specifically,
the question tokens are given as input to the \name~and interact with the queries via the self-attention layers,
which can guide the \name's~cross-attention layers to focus on more informative image regions.


Following BLIP,
our VQA data includes the training and validation splits from VQAv2, as well as training samples from Visual Genome.
Table~\ref{tbl:vqa_finetune} demonstrates the state-of-the-art results of BLIP-2 among open-ended generation models.


\subsection{Image-Text Retrieval}
\input{table/retrieval}
Since image-text retrieval does not involve language generation,
we directly finetune the first-stage-pretrained model w/o LLM.
Specifically,
we finetune the image encoder together with \name~on COCO using the same objectives (\ie~ITC, ITM, and ITG) as pre-training.
We then evaluate the model for both image-to-text retrieval and text-to-image retrieval on COCO and Flickr30K~\cite{flickr} datasets.
During inference,
we follow~\citet{ALBEF,blip} which first select $k=128$ candidates based on the image-text feature similarity, followed by a re-ranking based on pairwise ITM scores.
We experiment with both ViT-L and ViT-g as the image encoder.
Detailed hyperparameters can be found in the appendix.

The results are shown in Table~\ref{tbl:retrieval}.
BLIP-2 achieves state-of-the-art performance with significant improvement over existing methods on zero-shot image-text retrieval.

\input{table/retrieval_ablation.tex}
The ITC and ITM losses are essential for image-text retrieval as they directly learn image-text similarity.
In Table~\ref{tbl:retrieval_ablation},
we show that the ITG (image-grounded text generation) loss is also beneficial for image-text retrieval.
This result supports our intuition in designing the representation learning objectives: the ITG loss enforces the queries to extract visual features most relevant to the text,
thus improving vision-language alignment.






%% file: table/zeroshot.tex
\begin{table*}[!t]
	\centering	
    \setlength\tabcolsep{5pt}
	\resizebox{1\textwidth}{!}{%
	\begin{tabular}	{l  l l|  c  c  c c c }
		\toprule	 	
	 \multirow{3}{*}{Models}&\multirow{3}{*}{\makecell[l]{\#Trainable \\ Params}} &\multirow{3}{*}{\makecell[l]{Open-\\ sourced?}} &  \textbf{Visual Question Answering} & \multicolumn{2}{c}{\textbf{Image Captioning}}  & \multicolumn{2}{c}{\textbf{Image-Text Retrieval}} \\
	 & & & VQAv2 (test-dev) & \multicolumn{2}{c}{NoCaps (val)}  & \multicolumn{2}{c}{Flickr (test)} \\
	 & & & VQA acc. & CIDEr & SPICE & TR@1 & IR@1 \\
		\midrule
		BLIP~\cite{blip} & 583M & \checkmark & - & 113.2 & 14.8 & 96.7 & 86.7 \\
		SimVLM~\cite{simvlm} & 1.4B& \xmark & - & 112.2 & - & - & - \\
		BEIT-3~\cite{beit3} & 1.9B& \xmark & - & - & - & 94.9 & 81.5 \\
		Flamingo~\cite{flamingo} & 10.2B& \xmark & 56.3 & - & - & - & - \\
		\midrule
		BLIP-2  & 188M& \checkmark & \textbf{65.0} &  \textbf{121.6}& \textbf{15.8}  & \textbf{97.6} & \textbf{89.7}\\
		\bottomrule	
	\end{tabular}
 	}
 \vspace{-2ex}
	\caption
	{
	\small	
	    Overview of BLIP-2 results on various \textbf{zero-shot} vision-language tasks. Compared with previous state-of-the-art models.
     BLIP-2 achieves the highest zero-shot performance while requiring the least number of trainable parameters during vision-language pre-training.
	}
	\label{tbl:zeroshot}
\end{table*}	

%% file: table/vqa_zeroshot.tex
\begin{table*}[!t]
	\centering	
	\resizebox{0.75\textwidth}{!}{%
	\begin{tabular}	{l  l l |  c  c  c   c  }
		\toprule	 	
	 \multirow{2}{*}{Models}&\multirow{2}{*}{\makecell[l]{\#Trainable \\ Params}}&\multirow{2}{*}{\makecell[l]{\#Total \\ Params}} &  \multicolumn{2}{c}{VQAv2} & OK-VQA  & GQA \\
	 & & & val & test-dev & test & test-dev \\
		\midrule
		VL-T5$_\text{no-vqa}$ & 224M & 269M &13.5 & - & 5.8 &  6.3 \\
		FewVLM~\cite{FewVLM} & 740M & 785M & 47.7 & - & 16.5  & 29.3 \\		
		Frozen~\cite{Frozen} & 40M & 7.1B& 29.6 & - & 5.9 & - \\

		VLKD~\cite{VLKD} & 406M & 832M & 42.6 & 44.5 & 13.3 & -\\
		Flamingo3B~\cite{flamingo}&1.4B & 3.2B  & - & 49.2 & 41.2  & - \\
		Flamingo9B~\cite{flamingo}& 1.8B & 9.3B  & - & 51.8& 44.7& - \\
		Flamingo80B~\cite{flamingo}& 10.2B & 80B & - & 56.3 & \textbf{50.6} & - \\
		\midrule
		BLIP-2 ViT-L OPT$_\text{2.7B}$ & 104M& 3.1B   &  50.1 & 49.7  & 30.2 & 33.9 \\
  	BLIP-2 ViT-g OPT$_\text{2.7B}$ & 107M& 3.8B & 53.5 & 52.3 & 31.7 &  34.6\\
	BLIP-2 ViT-g OPT$_\text{6.7B}$ & 108M& 7.8B & 54.3 & 52.6 & 36.4 &  36.4\\
             BLIP-2 ViT-L FlanT5$_\text{XL}$& 103M& 3.4B  & 62.6 & 62.3   & 39.4 & \underline{44.4} \\
		BLIP-2 ViT-g FlanT5$_\text{XL}$& 107M& 4.1B  & \underline{63.1} &  \underline{63.0}  & 40.7  & 44.2 \\		
		BLIP-2 ViT-g FlanT5$_\text{XXL}$ & 108M & 12.1B & \textbf{65.2}  & \textbf{65.0} & \underline{45.9}  & \textbf{44.7} \\			
		\bottomrule	
	\end{tabular}
 	}
 \vspace{-1ex}
	\caption
	{
	\small	
		Comparison with state-of-the-art methods on zero-shot visual question answering.
	}
	\vspace{-1.5ex}
	\label{tbl:vqa_zeroshot}
\end{table*}		 


%% file: figure/qformer_effect.tex
\begin{figure}[!t]
 \centering
 \small
 \begin{minipage}{0.49\columnwidth}
	\centering
	\includegraphics[trim=0 0 5 0,clip,width=\textwidth]{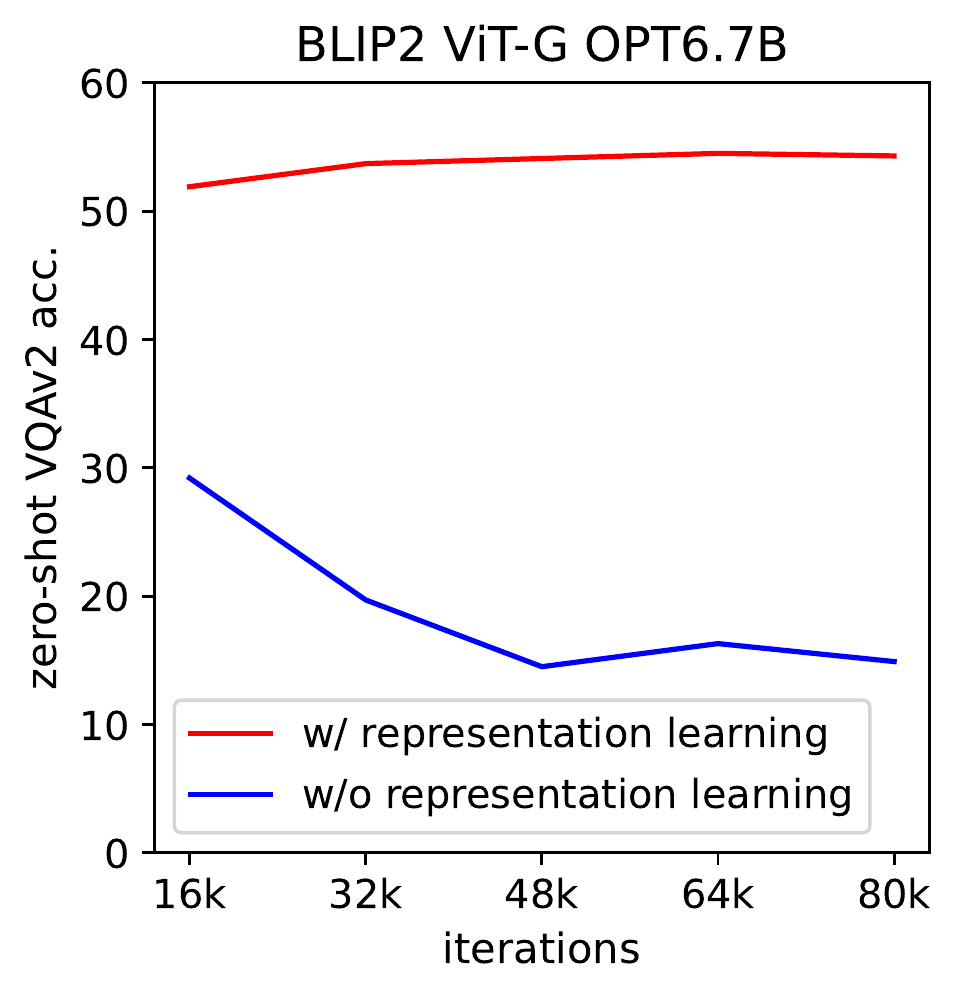}
\end{minipage}
\hfill
 \begin{minipage}{0.49\columnwidth}
	\centering
	\includegraphics[trim=0 0 5 0,clip,width=\textwidth]{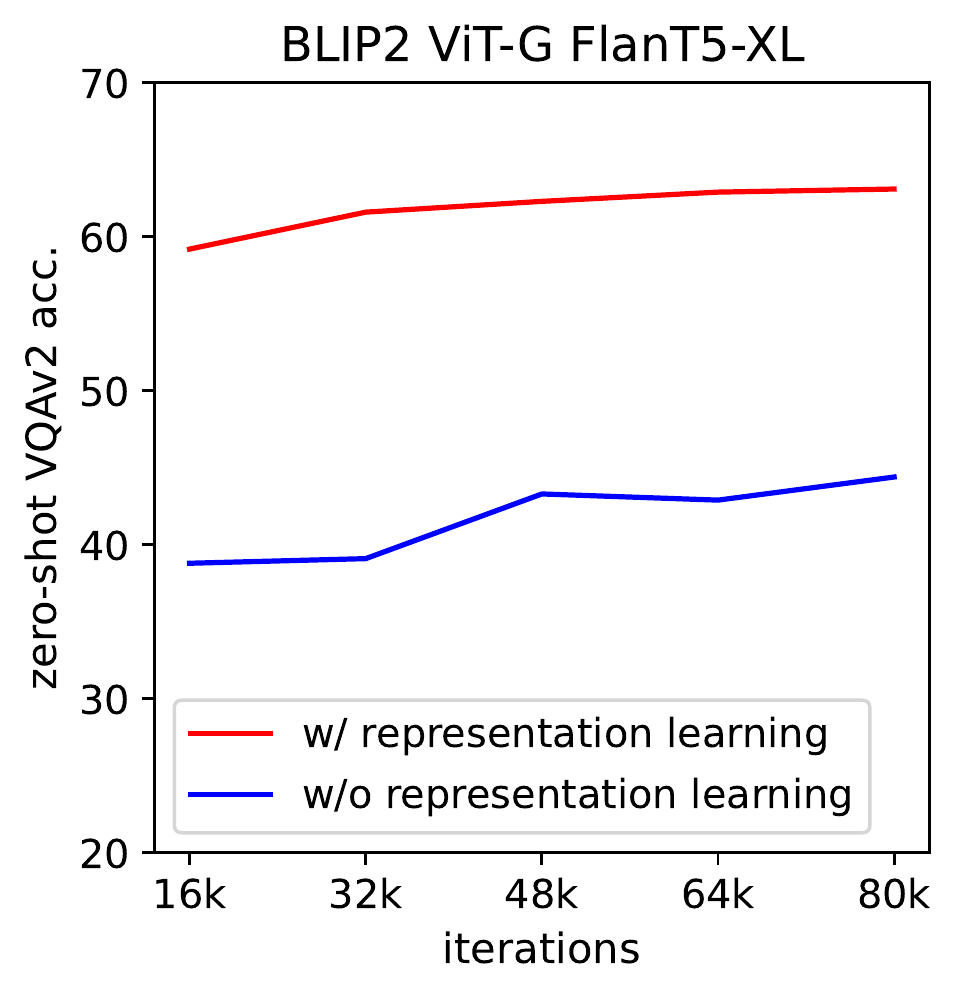}
\end{minipage}
\vspace{-1.5ex}
  \caption
  	{  	\small
  		Effect of vision-language representation learning on vision-to-language generative learning. Without representation learning, the \name~fails the bridge the modality gap, leading to significantly lower performance on zero-shot VQA.
	  } 
  \vspace{-2ex}
  \label{fig:qformer_effect}
 \end{figure}

%% file: table/caption.tex
\begin{table*}[!t]
	\centering	
	\resizebox{1\textwidth}{!}{%
	\begin{tabular}	{l  l |  c  c  c  c  c  c  c  c | c  c }
		\toprule	 	
	 \multirow{3}{*}{Models}&\multirow{3}{*}{\makecell[l]{\#Trainable \\ Params}}&  \multicolumn{8}{c|}{NoCaps Zero-shot (validation set)} & 
	 \multicolumn{2}{c}{COCO Fine-tuned}\\
	&  & \multicolumn{2}{c}{in-domain} & \multicolumn{2}{c}{near-domain} & \multicolumn{2}{c}{out-domain} & \multicolumn{2}{c|}{overall} & \multicolumn{2}{c}{Karpathy test}\\
	&   & C & S & C & S & C & S & C & S & B@4 & C\\
	  \midrule
	   OSCAR~\cite{oscar} & 345M & - & - & - & - & - & - & 80.9 & 11.3 & 37.4 & 127.8 \\
	   VinVL~\cite{vinvl} & 345M  & 103.1 & 14.2 & 96.1 & 13.8 & 88.3 & 12.1 & 95.5 & 13.5 &  38.2 & 129.3 \\
	   BLIP~\cite{blip} & 446M & 114.9 & 15.2 & 112.1 & 14.9 & 115.3 & 14.4 & 113.2 & 14.8 & 40.4 & 136.7 \\	 
	    OFA~\cite{ofa} & 930M & - & - & - & -& - & -& -& -& \textbf{43.9}& \underline{145.3}\\	   
	   
	   Flamingo~\cite{flamingo} & 10.6B & - & - & - & -& - & -& -& -& -&138.1\\
 SimVLM~\cite{simvlm} & $\sim$1.4B  &  113.7 &  - &  110.9 &  - &  115.2 &  - &  112.2 &  - &  40.6 &  143.3\\	   	   
	    \midrule
	    BLIP-2 ViT-g OPT$_\text{2.7B}$ & 1.1B  & \underline{123.0} & \underline{15.8} & 117.8 & \underline{15.4} & 123.4 & \textbf{15.1} & 119.7 & \underline{15.4} & \underline{43.7} & \textbf{145.8} \\		
		BLIP-2 ViT-g OPT$_\text{6.7B}$ & 1.1B  & \textbf{123.7} & \underline{15.8} & \underline{119.2} & 15.3 & \underline{124.4} & 14.8 & \underline{121.0} & 15.3 & 43.5 & 145.2 \\
	    BLIP-2 ViT-g FlanT5$_\text{XL}$ &1.1B  & \textbf{123.7} & \textbf{16.3} & \textbf{120.2} & \textbf{15.9}& \textbf{124.8} & \textbf{15.1}& \textbf{121.6}& \textbf{15.8}& 42.4 & 144.5\\  
		\bottomrule
  
	\end{tabular}
 	}
  \vspace{-2ex}
	\caption
	{
	\small	
		Comparison with state-of-the-art image captioning methods on NoCaps and COCO Caption.
		All methods optimize the cross-entropy loss during finetuning. C: CIDEr, S: SPICE, B@4: BLEU@4.	
	}
 \vspace{-1ex}
	\label{tbl:caption}
\end{table*}

%% file: table/vqa_finetune.tex
\begin{table}[!t]
	\centering	
    \setlength\tabcolsep{4pt}
	\resizebox{1\columnwidth}{!}{%
	\begin{tabular}	{l  l |  c  c  }
		\toprule	 	
	 \multirow{2}{*}{Models}&\multirow{2}{*}{\makecell[l]{\#Trainable \\ Params}}&  \multicolumn{2}{c}{VQAv2} \\
	 & & test-dev & test-std\\
		\midrule
		\multicolumn{3}{l}{~~\textit{Open-ended generation models}}\\
        ALBEF~\cite{ALBEF} & 314M  & 75.84 & 76.04\\
        BLIP~\cite{blip} & 385M  & 78.25 & 78.32 \\
        OFA~\cite{ofa} & 930M  & 82.00 & 82.00 \\
		Flamingo80B~\cite{flamingo} & 10.6B & 82.00 & 82.10 \\
       \textbf{BLIP-2} ViT-g FlanT5$_\text{XL}$ &1.2B & 81.55 & 81.66\\
       \textbf{BLIP-2} ViT-g OPT$_\text{2.7B}$ & 1.2B & 81.59& 81.74 
       \\
       \textbf{BLIP-2} ViT-g OPT$_\text{6.7B}$ & 1.2B & \textbf{82.19} & \textbf{82.30}
       \\
		\midrule
		\multicolumn{3}{l}{~~\textit{Closed-ended classification models}}\\
		
		VinVL & 345M  & 76.52 & 76.60 \\
		SimVLM~\cite{simvlm} & $\sim$1.4B  & 80.03 & 80.34 \\
		CoCa~\cite{coca} & 2.1B  & 82.30 & 82.30 \\
		BEIT-3~\cite{beit3} & 1.9B & \textbf{84.19} & \textbf{84.03} \\
	\bottomrule	
	\end{tabular}
 	}
 \vspace{-1ex}
	\caption
	{
	\small	
		Comparison with state-of-the-art models fine-tuned for visual question answering.
	}
	\vspace{-2ex}
	\label{tbl:vqa_finetune}
\end{table}

%% file: table/retrieval.tex
\begin{table*}[!t]
    \small
	\centering	
 \setlength\tabcolsep{4pt}
	\resizebox{\textwidth}{!}{%
	\begin{tabular}	{l l |  c  c  c  c   c  c | c  c  c  c  c  c }
		\toprule	 	
	 \multirow{3}{*}{Model} & \multirow{3}{*}{\makecell[l]{\#Trainable \\ Params}} &\multicolumn{6}{c|}{Flickr30K Zero-shot (1K test set)}   & \multicolumn{6}{c}{COCO Fine-tuned (5K test set)}\\
	 & &    \multicolumn{3}{c}{Image $\rightarrow$ Text}& \multicolumn{3}{c|}{Text $\rightarrow$ Image} &  \multicolumn{3}{c}{Image $\rightarrow$ Text}& \multicolumn{3}{c}{Text $\rightarrow$ Image}\\
	     
	& &  R@1 &R@5 &R@10& R@1 &R@5&R@10& R@1 &R@5&R@10& R@1 &R@5&R@10\\
	\midrule  
    \textit{~~Dual-encoder models} \\
    CLIP~\cite{clip} & 428M &
	88.0 & 98.7 & 99.4 & 68.7 & 90.6 & 95.2 
	& - & - &-& - & - &- 
	\\    
	ALIGN~\cite{align} & 820M 
	& 88.6 &  98.7 &99.7& 75.7 & 93.8 &96.8
	& 77.0 & 93.5 &96.9&  59.9 & 83.3 &89.8 
	\\    
	FILIP~\cite{filip} &417M 
	&89.8& 99.2& 99.8 &75.0 &93.4& 96.3
	& 78.9& 94.4& 97.4& 61.2& 84.3& 90.6
	\\
	Florence~\cite{florence}& 893M 
	&90.9& 99.1 &- &76.7& 93.6& -
	& 81.8 & 95.2 & - & 63.2 & 85.7 & -
	\\
	BEIT-3\cite{beit3} &1.9B 
	&94.9 & 99.9&\textbf{100.0} & 81.5 & 95.6 &97.8 
	& \underline{84.8} & \underline{96.5}&\underline{98.3} & \underline{67.2}& \textbf{87.7} & \textbf{92.8}
	\\
	 \midrule
    \textit{~~Fusion-encoder models} \\	   
	UNITER~\cite{uniter} & 303M
	& 83.6 & 95.7 & 97.7 & 68.7 & 89.2 & 93.9 
	&65.7  &88.6 &93.8 &52.9 &79.9 &88.0 
	\\
	OSCAR~\cite{oscar} & 345M 
	& - & - & - & - & - & - 
	& 70.0 & 91.1 &95.5 &  54.0 & 80.8& 88.5 
	\\
	VinVL~\cite{vinvl} & 345M
	& - & - & - & - & - & - 
	& 75.4 & 92.9&  96.2&  58.8&  83.5&  90.3
	\\	
	\midrule
	\multicolumn{2}{l}{\textit{~~Dual encoder + Fusion encoder reranking}} \\ 
	ALBEF~\cite{ALBEF} &233M 
	& 94.1 & 99.5 &99.7 & 82.8 & 96.3 &98.1
	&77.6 & 94.3 &97.2 & 60.7& 84.3 &90.5  
	\\
    BLIP~\cite{blip} &446M  
    & 96.7 & \textbf{100.0}& \textbf{100.0} & 86.7 & 97.3 & 98.7
    & 82.4 & 95.4 &97.9& 65.1 & 86.3 &91.8 
    \\    
    \textbf{BLIP-2} ViT-L & 474M 
    & \underline{96.9} & \textbf{100.0}& \textbf{100.0} & \underline{88.6} & \underline{97.6} & \textbf{98.9}
    & 83.5 & 96.0 &98.0& 66.3 & 86.5 &91.8 \\
    
    \textbf{BLIP-2} ViT-g & 1.2B
     & \textbf{97.6} & \textbf{100.0}&\textbf{100.0}  &  \textbf{89.7}& \textbf{98.1} &\textbf{98.9}
    & \textbf{85.4} &  \textbf{97.0} &\textbf{98.5}& \textbf{68.3} & \textbf{87.7} &\underline{92.6}\\  
    \bottomrule
	\end{tabular}
	}
  \vspace{-2ex}
	\caption
	{
	\small	
        Comparison with state-of-the-art image-text retrieval methods, finetuned on COCO and zero-shot transferred to Flickr30K.
	}
	\label{tbl:retrieval}
 \vspace{-1ex}
\end{table*}

%% file: table/retrieval_ablation.tex
\begin{table}[!t]
   
	\centering	
  \resizebox{0.9\columnwidth}{!}{%
 \begin{tabular}	{l  |  c  c  c c }
        \toprule
        \multirow{2}{*}{\makecell[l]{COCO finetuning \\ objectives}} & \multicolumn{2}{c}{Image $\rightarrow$ Text} & \multicolumn{2}{c}{Text $\rightarrow$ Image} \\
        & R@1 & R@5  & R@1 & R@5 \\
        \midrule
        ITC + ITM & 84.5 & 96.2 & 67.2 & 87.1 \\
        ITC + ITM + ITG & 85.4 & 97.0 & 68.3 & 87.7 \\
 	\bottomrule	
	\end{tabular}
}
 \vspace{-1ex}
	\caption
	{
	\small	
        The image-grounded text generation (ITG) loss improves image-text retrieval performance by enforcing the queries to extract language-relevant visual features.
	}
	\vspace{-1ex}
	\label{tbl:retrieval_ablation}
\end{table}		 

%% file: 5-limitation.tex
\section{Limitation}
\label{sec:limitation}
\vspace{-0.5ex}

Recent LLMs can perform in-context learning given few-shot examples.
However,
our experiments with BLIP-2 do not observe an improved VQA performance when providing the LLM with in-context VQA examples.
We attribute the lack of in-context learning capability to our pre-training dataset,
which only contains a single image-text pair per sample.
The LLMs cannot learn from it the correlation among multiple image-text pairs in a single sequence.
The same observation is also reported in the Flamingo paper,
which uses a close-sourced interleaved image and text dataset (M3W) with multiple image-text pairs per sequence.
We aim to create a similar dataset in future work.

BLIP-2's image-to-text generation could have unsatisfactory results due to various reasons including inaccurate knowledge from the LLM,
activating the incorrect reasoning path,
or not having up-to-date information about new image content (see Figure~\ref{fig:example_limitation}).
Furthermore,
due to the use of frozen models,
BLIP-2 inherits the risks of LLMs,
such as outputting offensive language, propagating social bias, or leaking private information.
Remediation approaches include using instructions to guide model's generation or training on a filtered dataset with harmful content removed.


\vspace{-0.5ex}
\section{Conclusion}
\vspace{-0.5ex}
\label{sec:conclusion}
We propose BLIP-2,
a generic and compute-efficient method for vision-language pre-training that leverages frozen pre-trained image encoders and LLMs.
BLIP-2 achieves state-of-the-art performance on various vision-language tasks while having a small amount of trainable parameters during pre-training.
BLIP-2 also demonstrates emerging capabilities in zero-shot instructed image-to-text generation.
We consider BLIP-2 as an important step towards building a multimodal conversational AI agent.

%% file: appendix.tex
\newpage
\clearpage

\appendix

\begin{table*}[!t]
    \small
	\centering	
 \begin{tabular}	{l  |  c  c  c  }
        \toprule
        LLM & FlanT5$_\text{XL}$ & OPT$_\text{2.7B}$ & OPT$_\text{6.7B}$\\
        \midrule
        Fine-tuning epochs & \multicolumn{3}{c}{5}\\
        Warmup steps & \multicolumn{3}{c}{1000}\\
        Learning rate & \multicolumn{3}{c}{1e-5} \\
        Batch size & \multicolumn{3}{c}{256} \\
        AdamW $\beta$ & \multicolumn{3}{c}{(0.9,0.999)} \\
        Weight decay & \multicolumn{3}{c}{0.05}\\
        Drop path & \multicolumn{3}{c}{0}\\
        Image resolution & \multicolumn{3}{c}{364}\\
        Prompt & \multicolumn{3}{c}{``a photo of''}\\
        Inference beam size & \multicolumn{3}{c}{5}\\   
        Layer-wise learning rate decay for ViT & 1 & 1 & 0.95\\
 	\bottomrule	
	\end{tabular}

 \vspace{-1ex}
	\caption
	{
	\small	
        Hyperparameters for fine-tuning BLIP-2 with ViT-g on COCO captioning.
	}
	\vspace{-1ex}
\end{table*}

\begin{table*}[!t]
    \small
	\centering	
 \begin{tabular}	{l  |  c  c  c  }
        \toprule
        LLM & FlanT5$_\text{XL}$ & OPT$_\text{2.7B}$ & OPT$_\text{6.7B}$\\
        \midrule
        Fine-tuning epochs & \multicolumn{3}{c}{5}\\
        Warmup steps & \multicolumn{3}{c}{1000}\\
        Learning rate & \multicolumn{3}{c}{1e-5} \\
        Batch size & \multicolumn{3}{c}{128} \\
        AdamW $\beta$ & \multicolumn{3}{c}{(0.9,0.999)} \\
        Weight decay & \multicolumn{3}{c}{0.05}\\
        Drop path & \multicolumn{3}{c}{0}\\
        Image resolution & \multicolumn{3}{c}{490}\\
        Prompt & \multicolumn{3}{c}{``Question: \{\} Answer:''}\\
        Inference beam size & \multicolumn{3}{c}{5}\\   
        Layer-wise learning rate decay for ViT & 0.95 & 0.95 & 0.9\\
 	\bottomrule	
	\end{tabular}

 \vspace{-1ex}
	\caption
	{
	\small	
        Hyperparameters for fine-tuning BLIP-2 with ViT-g on VQA.
	}
	\vspace{-1ex}
\end{table*}

\begin{table*}[!t]
    \small
	\centering	
 \begin{tabular}	{l  |  c  c  }
        \toprule
        Image Encoder & ViT-L/14 & ViT-g/14\\
        \midrule
        Fine-tuning epochs & \multicolumn{2}{c}{5}\\
        Warmup steps & \multicolumn{2}{c}{1000}\\
        Learning rate& 5e-6 & 1e-5  \\
        Batch size & \multicolumn{2}{c}{224} \\
        AdamW $\beta$ & (0.9,0.98) & (0.9,0.999) \\
        Weight decay & \multicolumn{2}{c}{0.05}\\
        Drop path & \multicolumn{2}{c}{0}\\
        Image resolution & \multicolumn{2}{c}{364}\\
        Layer-wise learning rate decay for ViT & 1 & 0.95\\
 	\bottomrule	
	\end{tabular}

 \vspace{-1ex}
	\caption
	{
	\small	
        Hyperparameters for fine-tuning BLIP-2 on COCO image-text retrieval.
	}
	\vspace{-1ex}
\end{table*}

\begin{figure*}[b]
\centering
  \includegraphics[width=\textwidth]{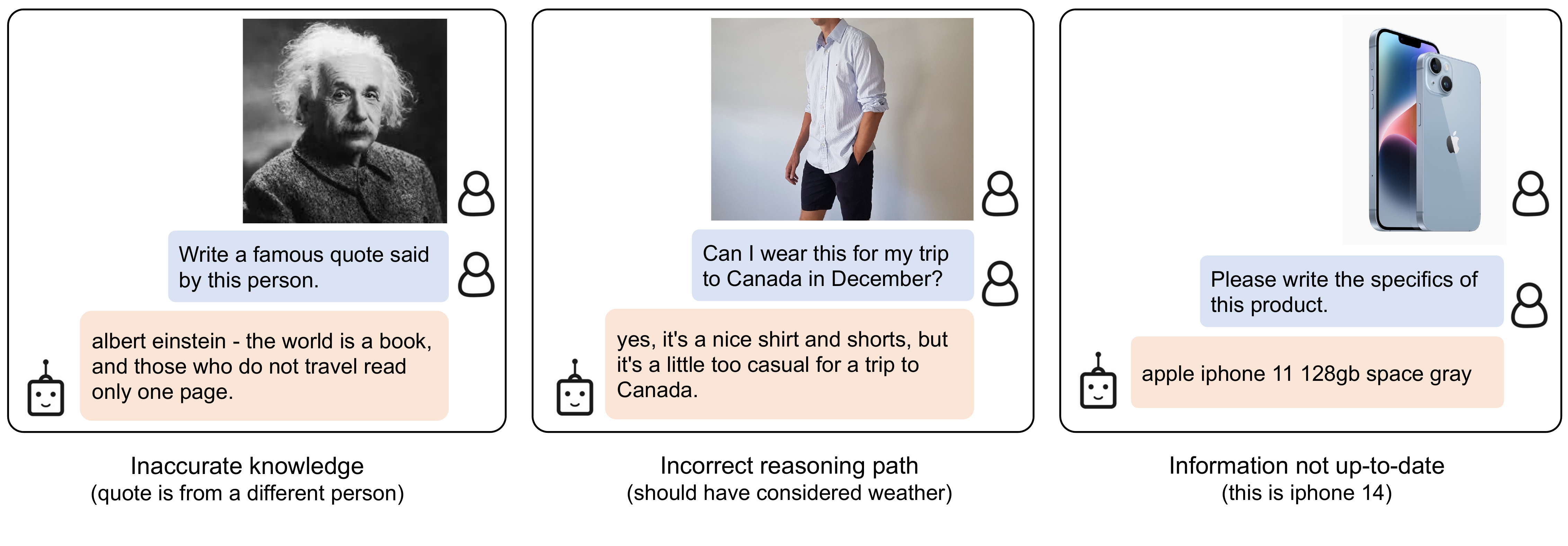}
\vspace{-6ex}
\caption{Incorrect output examples for instructed zero-shot image-to-text generation using a BLIP-2 model w/ ViT-g and FlanT5$_\text{XXL}$.
}
\clearpage
\label{fig:example_limitation}
\end{figure*}

\begin{figure}[t!]
\centering
  \includegraphics[width=0.48\textwidth]{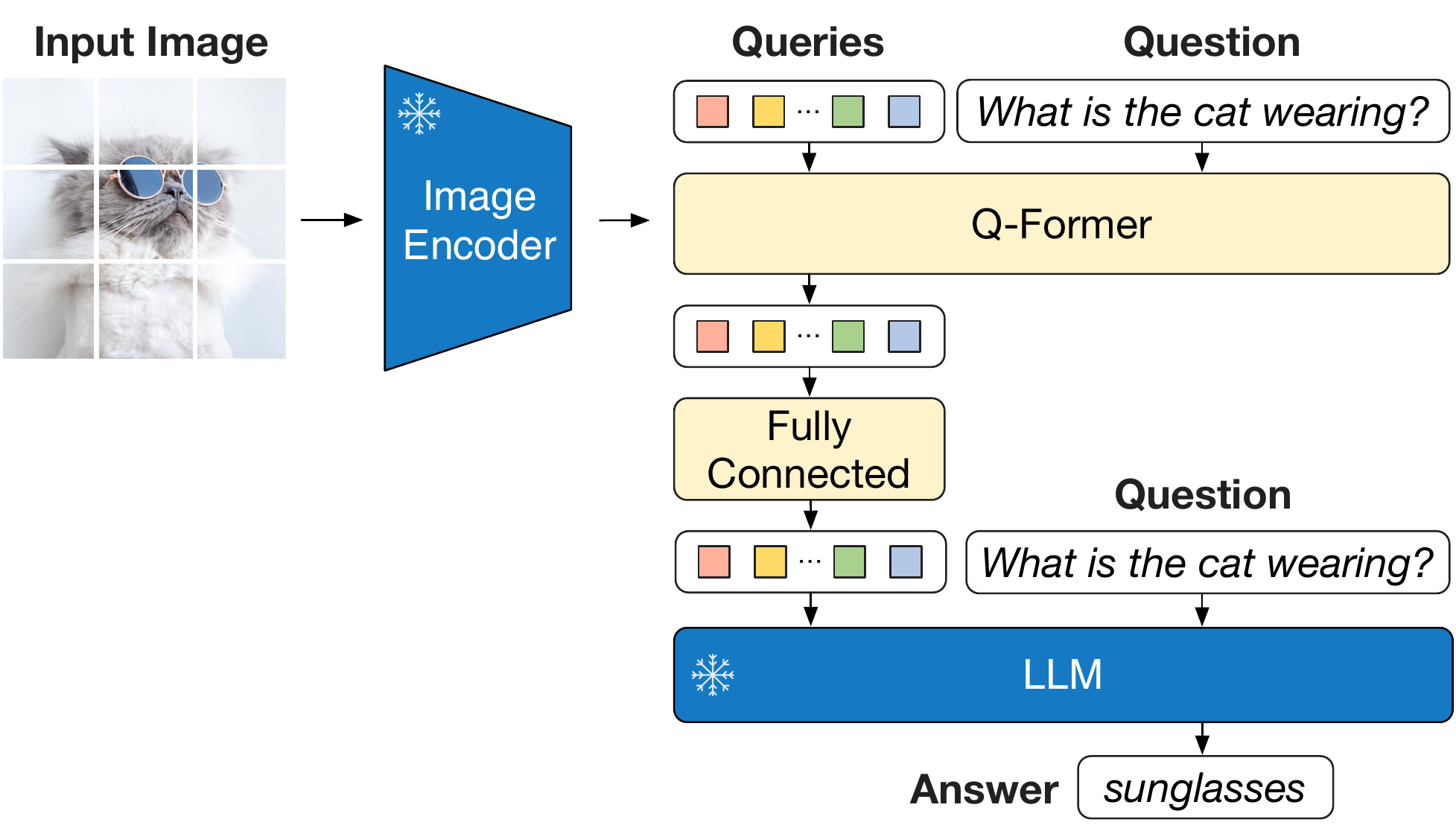}
\caption{Model architecture for VQA finetuning, where the LLM receives Q-Former's output and the question as input, then predicts answers. We also provide the question as a condition to Q-Former, such that the extracted image features are more relevant to the question.
}
\label{fig:example_limitation}
\end{figure}